%% file: main_file.tex
\documentclass[sigconf]{acmart}
\AtBeginDocument{%
  \providecommand\BibTeX{{%
    \normalfont B\kern-0.5em{\scshape i\kern-0.25em b}\kern-0.8em\TeX}}}

\copyrightyear{2024}
\acmYear{2024}
\setcopyright{acmlicensed}\acmConference[KDD '24]{Proceedings of the 30th ACM SIGKDD Conference on Knowledge Discovery and Data Mining}{August 25--29, 2024}{Barcelona, Spain}
\acmBooktitle{Proceedings of the 30th ACM SIGKDD Conference on Knowledge Discovery and Data Mining (KDD '24), August 25--29, 2024, Barcelona, Spain}
\acmDOI{10.1145/3637528.3671561}
\acmISBN{979-8-4007-0490-1/24/08}

\usepackage{makecell}
\usepackage{multirow}
\usepackage{algorithm}
\usepackage{algpseudocode}
\usepackage{tabularx}
\usepackage{graphicx}
\usepackage{enumitem}
\usepackage{multicol}
\newcommand{\systemname}{\textit{LiRank}}
\newcounter{savealgorithm}

\makeatother

\algnewcommand\algorithmicforeach{\textbf{for each}}
\algdef{S}[FOR]{ForEach}[1]{\algorithmicforeach\ #1\ \algorithmicdo}

\algdef{SE}[REPEATN]{RepeatN}{End}[1]{\algorithmicrepeat\ #1 \textbf{times}}{\algorithmicend}

\begin{document}

\title{LiRank: Industrial Large Scale Ranking Models at LinkedIn}


\author{Fedor Borisyuk}
\author{Mingzhou Zhou}
\author{Qingquan Song}
\author{Siyu Zhu}
\affiliation{
 \institution{LinkedIn}
 \city{Mountain View}
 \state{CA}
 \country{USA}}
\email{fedorvb@gmail.com}

\author{Birjodh Tiwana}
\author{Ganesh Parameswaran}
\author{Siddharth Dangi}
\author{Lars Hertel}
\affiliation{
 \institution{LinkedIn}
 \city{Mountain View}
 \state{CA}
 \country{USA}}
\email{btiwana@linkedin.com}

\author{Qiang Charles Xiao}
\author{Xiaochen Hou}
\author{Yunbo Ouyang}
\author{Aman Gupta}
\affiliation{
 \institution{LinkedIn}
 \city{Mountain View}
 \state{CA}
 \country{USA}}
\email{cxiao.uoft@gmail.com}

\author{Sheallika Singh}
\author{Dan Liu}
\author{Hailing Cheng}
\author{Lei Le}
\affiliation{
 \institution{LinkedIn}
 \city{Mountain View}
 \state{CA}
 \country{USA}}
\email{sheasingh@linkedin.com}

\author{Jonathan Hung}
\author{Sathiya Keerthi}
\author{Ruoyan Wang}
\author{Fengyu Zhang}
\affiliation{
 \institution{LinkedIn}
 \city{Mountain View}
 \state{CA}
 \country{USA}}
\email{jhung@linkedin.com}

\author{Mohit Kothari}
\author{Chen Zhu}
\author{Daqi Sun}
\author{Yun Dai}
\affiliation{
 \institution{LinkedIn}
 \city{Mountain View}
 \state{CA}
 \country{USA}}
\email{mkothari@linkedin.com}

\author{Xun Luan}
\author{Sirou Zhu}
\author{Zhiwei Wang}
\author{Neil Daftary}
\affiliation{
 \institution{LinkedIn}
 \city{Mountain View}
 \state{CA}
 \country{USA}}
\email{xun.luan.rice@gmail.com}

\author{Qianqi Shen}
\author{Chengming Jiang}
\author{Haichao Wei}
\author{Maneesh Varshney}
\author{Amol Ghoting}
\author{Souvik Ghosh}
\affiliation{
 \institution{LinkedIn}
 \city{Mountain View}
 \state{CA}
 \country{USA}}
 \email{qishen@linkedin.com}

\renewcommand{\shortauthors}{Fedor Borisyuk et al.}

\begin{abstract}
\input{abstract}

\end{abstract}



\maketitle
\section{Introduction}\label{sec:intro}
\input{intro.tex}

\input{related_work.tex}

\section{Large Ranking Models}\label{sec:overview}
\input{overview.tex}


\input{training_scalability.tex}

\input{Experiments.tex}


\input{deployment_lessons.tex}

\section{Conclusion}\label{sec:conclusion}
\input{conclusion.tex}


\bibliographystyle{ACM-Reference-Format}
\balance
\bibliography{bibliography}

\input{reproducability.tex}

\end{document}

%% file: abstract.tex
We present {\systemname}, a large-scale ranking framework at LinkedIn that brings to production state-of-the-art modeling architectures and optimization methods. We unveil several modeling improvements, including Residual DCN, which adds attention and residual connections to the famous DCNv2 architecture.  We share insights into combining and tuning SOTA architectures to create a unified model, including Dense Gating, Transformers and Residual DCN. We also propose novel techniques for calibration and describe how we productionalized deep learning based explore/exploit methods.

To enable effective, production-grade serving of large ranking models, we detail how to train and compress models using quantization and vocabulary compression. We provide details about the deployment setup for large-scale use cases of Feed ranking, Jobs Recommendations, and Ads click-through rate (CTR) prediction.

We summarize our learnings from various A/B tests by elucidating the most effective technical approaches. These ideas have contributed to relative metrics improvements across the board at LinkedIn: +0.5\% member sessions in the Feed, +1.76\% qualified job applications for Jobs search and recommendations, and +4.3\% for Ads CTR. We hope this work can provide practical insights and solutions for practitioners interested in leveraging large-scale deep ranking systems.

%% file: intro.tex
LinkedIn is the world's largest professionals network with more than 1 billion members in more than 200 countries and territories worldwide. Hundreds of millions of LinkedIn members engage on a regular basis to find opportunities and connect with other professionals.

At LinkedIn, we strive to provide our members with valuable content that can help them build professional networks, learn new skills, and discover exciting job opportunities. To ensure this content is engaging and relevant, we aim to understand each member's specific preferences. This may include interests such as keeping up with the latest news and industry trends, participating in discussions by commenting or reacting, contributing to collaborative articles, sharing career updates, learning about new business opportunities, or applying for jobs.

In this paper, we introduce a set of innovative enhancements to model architectures and optimization strategies, all aimed at enhancing the member experience. The \textbf{contribution of the paper} consists of:
\begin{itemize}[leftmargin=*]
    \item We propose a novel Residual DCN layer (\S\ref{sec:residual_dcn}), an improvement on top of DCNv2\cite{Wang2017DeepC}, with attention and residual connections.
    \item We propose a novel isotonic calibration layer trained jointly within deep learning model (\S\ref{sec:isotonic_layer}).
    \item We provide customizations of deep-learning based exploit/explore methods to production (\S\ref{sec:explore_exploit}). 
    \item Integrating various architectures into a large-scale unified ranking model presented challenges such as diminishing returns (first attempt lead to no gain), overfitting, divergence, and different gains across applications. In \S\ref{sec:overview}, we discuss our approach to developing high-performing production ranking models, combining Residual DCN (\S\ref{sec:residual_dcn}), isotonic calibration layer (\S\ref{sec:isotonic_layer}), dense gating with larger MLP (\S\ref{sec:dense_gating}), incremental training (\S\ref{sec:incremental_training}), transformer-based history modeling (\S\ref{sec:trans_act}), deep learning explore-exploit strategies (\S\ref{sec:explore_exploit}), wide popularity features (\S\ref{sec:popularity_features}), multi-task learning (\S\ref{sec:multitask}),  dwell modeling (\S\ref{sec:dwell_modeling}).
    \item We share practical methods to speed up training process, enabling rapid model iteration (\S\ref{sec:training_scalability}).
    \item We provide insights into training and compressing deep ranking models using quantization (\S\ref{sec:model_quantization}) and vocabulary compression (\S\ref{sec:qr_and_murmur}) to facilitate the effective deployment of large-ranking models in production.
\end{itemize}

Proposed modeling advancements within this paper enabled our models to efficiently handle a larger number of parameters, leading to higher-quality content delivery. Within the paper we introduce details of large scale architectures of Feed ranking in \S\ref{sec:Overview:FeedRanking}, Ads CTR model \S\ref{sec:Overview:AdsCTRRanking}, and Job recommendation ranking models in \S\ref{sec:job_recommendations}.

In \S\ref{sec:experiments}, we detail our experiences in deploying large-ranking models in production for Feed Ranking, Jobs Recommendations, and Ads CTR prediction, summarizing key learnings gathered from offline experimentation and A/B tests. Notably, the techniques presented in this work have resulted in significant relative improvements: a 0.5\% increase in Feed sessions, a 1.76\% enhancement in the number of qualified applicants within Job Recommendations, and a 4.3\% boost in Ads CTR. We believe that this work can provide practical solutions and insights for engineers who are interested in applying large DNN ranking models at scale.

\vspace{-0.3em}

%% file: related_work.tex
\section{Related Work}\label{sec:related_work}
The use of deep neural network models in personalized recommender systems has been dominant in academia and industry since the success of the Wide\&Deep model\cite{wideDeep2016} in 2016. Typically, these models consist of feature embeddings, feature selection, and feature interaction components, with much research focused on enhancing feature interactions. The Wide\&Deep model\cite{wideDeep2016} initiated this trend by combining a generalized linear model with an MLP network. Subsequent research aimed to keep the MLP network for implicit feature interactions and replace the linear model with other modules for capturing explicit higher-order feature interactions. Examples include DeepFM\cite{Guo2017DeepFMAF}, which replaced the linear model with FM; deep cross network (DCN)\cite{Wang2017DeepC} and its follow-up DCNv2\cite{Wang2020DCNVI}, which introduced a cross network for high-order feature interactions; xDeepFM\cite{Lian2018xDeepFMCE}, offering compressed interaction network (CIN) for explicit vector-wise feature interactions; AutoInt\cite{Song2018AutoIntAF}, which introduced self-attention networks for explicit feature interaction; AFN\cite{Cheng2019AdaptiveFN}, exploring adaptive-order feature interactions through a logarithmic transformation layer; and FinalMLP\cite{Mao2023FinalMLPAE}, which achieved impressive performance by combining two MLPs.

We experimented with and customized these architectures for various LinkedIn recommender tasks, with DCNv2 proving to be the most versatile. We propose enhancements to DCNv2, referred to as Residual DCN, in this paper. Additionally, we implemented a model parallelism design in TensorFlow(TF), similar to the approach proposed in the DLRM\cite{Naumov2019DeepLR} paper, to accelerate model training with large embedding tables.

In our investigation, we've encountered challenges when attempting to seamlessly integrate original architectures into production environments. These challenges often manifest as issues such as model training divergence, over-fitting, or limited observable performance improvements. Crafting a high-performing model by effectively leveraging these architectures demands substantial effort, often characterized by a painstaking process of trial and error. Consequently, in this paper, we aim to offer valuable insights derived from our experiences in successfully assembling state-of-the-art (SOTA) architectures into production-ready ranking models.

While enhancing neural network predictive performance through various optimizations and architectures, the space of calibration remained relatively stable. Traditional industry-standard methods \cite{Calibration_overview} like Histogram binning, Platt Scaling, and Isotonic Regression are applied in post-processing steps after deep model training. Some research has introduced calibration-aware losses to address under/over calibration issues usually resulting in trade-off \cite{Calibration_soft_loss, ScaleCalibrationPaper} or slight improved metrics \cite{On_the_Factory_Floor}. In \S\ref{sec:isotonic_layer} we propose an isotonic calibration layer within the deep learning model which learns to calibrate deep model scores during model training and improves model predictive accuracy significantly.

%% file: overview.tex
In this section, we introduce large ranking models used by LinkedIn Feed Ranking and Ads CTR (click-through-rate) prediction. We observe that the choice of architecture components varies based on the use case. We'll share our insights on building effective ranking models for production scenarios.
\input{FeedRankingArch.tex}
\input{adsCTRArch.tex}

\input{attention_dcn.tex}

\input{calibration_architecture.tex}

\input{dense_gating.tex}


\input{incremental_training.tex}

\input{transact.tex}

\input{explore_exploit.tex}

\input{lpm_wide.tex}

\input{multi_task_learning.tex}

\input{dwell_modeling.tex}

\input{qr_and_murmur.tex}

\input{model_quantization.tex}

%% file: FeedRankingArch.tex
\subsection{Feed Ranking Model}\label{sec:Overview:FeedRanking}
The primary Feed ranking model employs a point-wise ranking approach, predicting multiple action probabilities including like, comment, share, vote, and long dwell and click for each <member, candidate post> pair. These predictions are linearly combined to generate the final post score. A TF model with a multi-task learning (MTL) architecture generates these probabilities in two towers: the click tower for probabilities of click and long dwell, and contribution tower for contribution and related predictions. Both towers use the same set of dense features normalized based on their distribution\cite{AirBnB_Search}, and apply multiple fully-connected layers. Sparse ID embedding features (\S\ref{sec:sparse_features}) are transformed into dense embeddings \cite{Naumov2019DeepLR} through lookup in embedding tables of Member/Actor and Hashtag Embedding Table as in Figure ~\ref{fig:FeedContributionTower}. For reproducability in appendix in Figure \ref{fig:feed_model_architecture_details} we provide a diagram showing how different architectures are connected together into a single model.

\begin{figure}
  \centering
  \includegraphics[height=5cm, width=7cm]{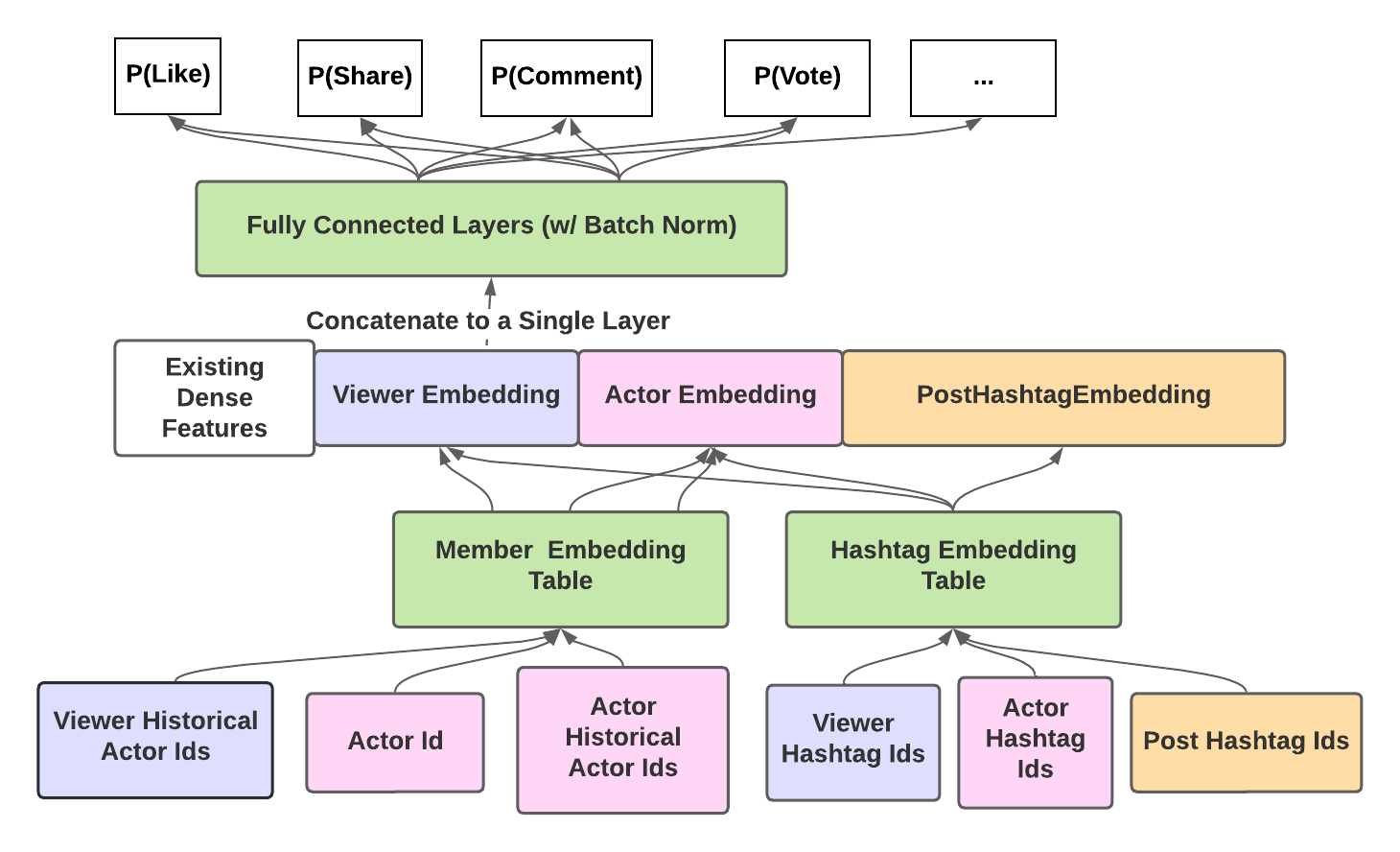}
  \caption{Contribution tower of the main Feed ranking model}
  \label{fig:FeedContributionTower}
\end{figure}

%% file: AdsCTRArch.tex
\subsection{Ads CTR Model}\label{sec:Overview:AdsCTRRanking}
\begin{figure}
 \centering
 \includegraphics[height=6cm]{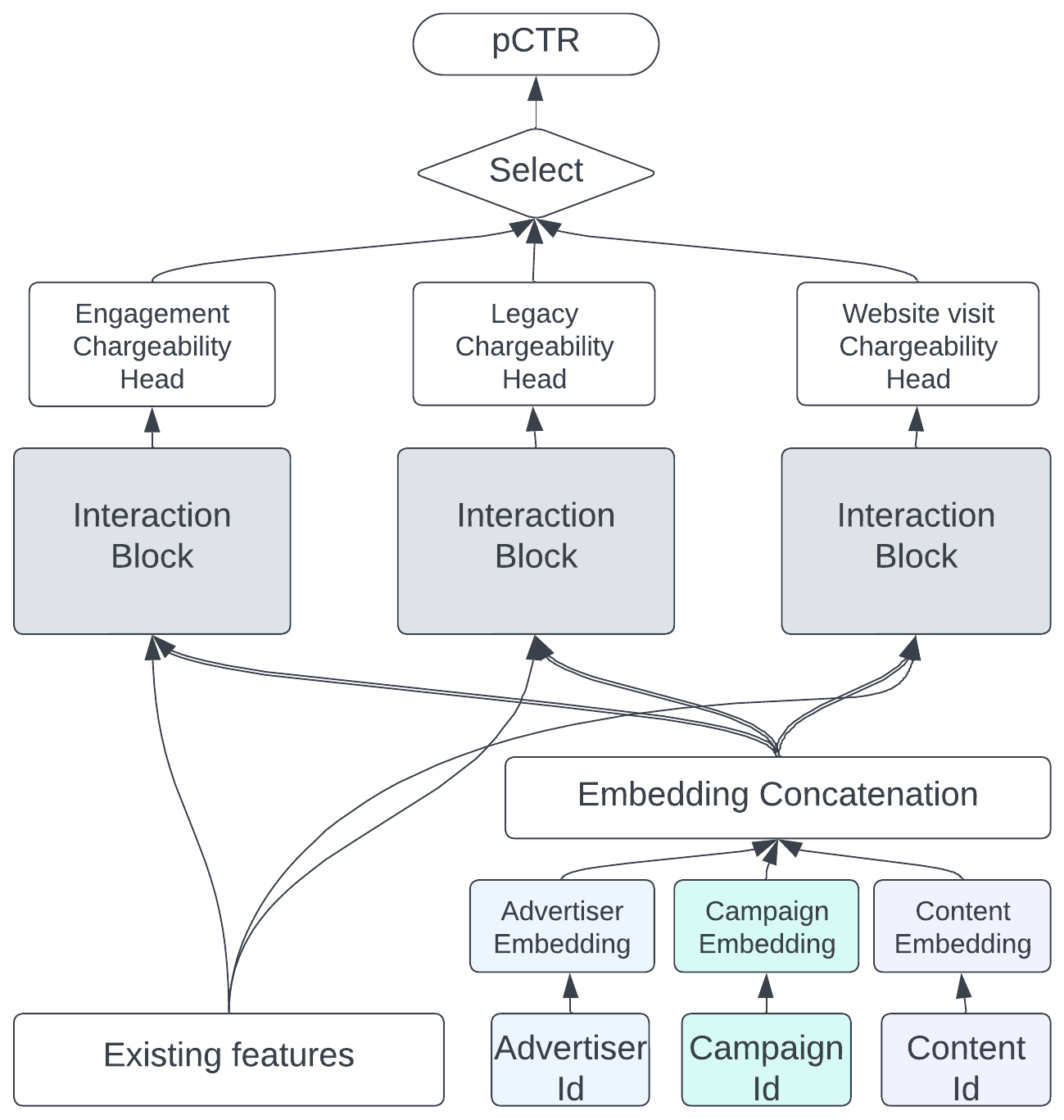}
 \caption{Ads CTR chargeability-based multi-task model}
 \label{fig:ads_ctr}
\end{figure}

At LinkedIn, ads selection relies on click-through-rate (CTR) prediction, estimating the likelihood of member clicks on recommended ads. This CTR probability informs ad auctions for displaying ads to members. Advertisers customize chargeable clicks for campaigns, such as some advertisers consider social interaction such as `like', `comment' as chargeable clicks while others only consider visiting ads websites as clicks. 
Usually only positive customized chargeable clicks are treated as positive labels. To better capture user interest, our CTR prediction model is a chargeability-based MTL model with 3 heads that correspond to 3 chargeability categorizations where similar chargeable definitions are grouped together regardless of advertiser customization. Each head employs independent interaction blocks such as MLP and DCNv2 blocks. The loss function combines head-specific losses. For features, besides traditional features from members and advertisers, we incorporate ID features to represent advertisers, campaigns, and advertisements. The model architecture is depicted in Figure ~\ref{fig:ads_ctr}.

%% file: attention_dcn.tex
\subsection{Residual DCN}\label{sec:residual_dcn}
\begin{figure}
  \centering
  \includegraphics[height=5cm,]{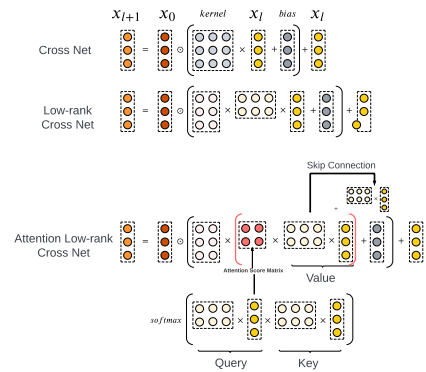}
  \caption{Residual Cross Network}
  \label{fig:improved_dcnv2}
\end{figure}
To automatically capture feature interactions, we utilized DCNv2 \cite{wang2021dcn}. Our offline experiments revealed that two DCNv2 layers provided sufficient interaction complexity, as adding more layers yielded diminishing relevance gains while increasing training and serving times significantly. Despite using just two layers, DCNv2 added a considerable number of parameters due to the large feature input dimension. To address this, we adopted two strategies for enhancing efficiency. First, following \cite{wang2021dcn}, we replaced the weight matrix with two skinny matrices resembling a low-rank approximation. Second, we reduced the input feature dimension by replacing sparse one-hot features with embedding-table look-ups, resulting in nearly a 30\% reduction. These modifications allowed us to substantially reduce DCNv2's parameter count with only minor effects on relevance gains, making it feasible to deploy the model on CPUs.

To further enhance the power of DCNv2, specifically, the cross-network, introduced an attention schema in the low-rank cross net. Specifically, the original low-rank mapping is duplicated as three with different mapping kernels, where the original one serves as the value matrix and the other two as the query and key matrices, respectively. An attention score matrix is computed and inserted between the low-rank mappings. Figure~\ref{fig:improved_dcnv2} describes the basic scaled dot-product self-attention. A temperature could also be added to balance the complicacy of the learned feature interactions. In the extreme case, the attention cross net will be degenerated to the normal cross net when the attention score matrix is an identity matrix. Practically, we find that adding a skip connection and fine-tuning the attention temperature is beneficial for helping learn more complicated feature correlations while maintain stable training. By paralleling a low-rank cross net with an attention low-rank cross net, we found a statistically significant improvement on feed ranking task (\S\ref{sec:feed_ablation}).
\vspace{-1.0em}

%% file: calibration_architecture.tex
\subsection{Isotonic Calibration Layer in DNN}\label{sec:isotonic_layer}
Model calibration ensures that estimated class probabilities align with real-world occurrences, a crucial aspect for business success. For example, Ads charging prices are linked to click-through rate (CTR) probabilities, making accurate calibration essential. It also enables fair comparisons between different models, as the model score distribution can change when using different models or objectives. Traditionally, calibration is performed post-training using classic methods like Platt scaling and isotonic regression. However, these methods are not well-suited for deep neural network models due to limitations like parameter space constraints and incompatibility. Additionally, scalability becomes challenging when incorporating multiple features like device, channel, or item IDs into calibration.

\begin{figure}[h]
\centering
\includegraphics[width=0.3\textwidth]{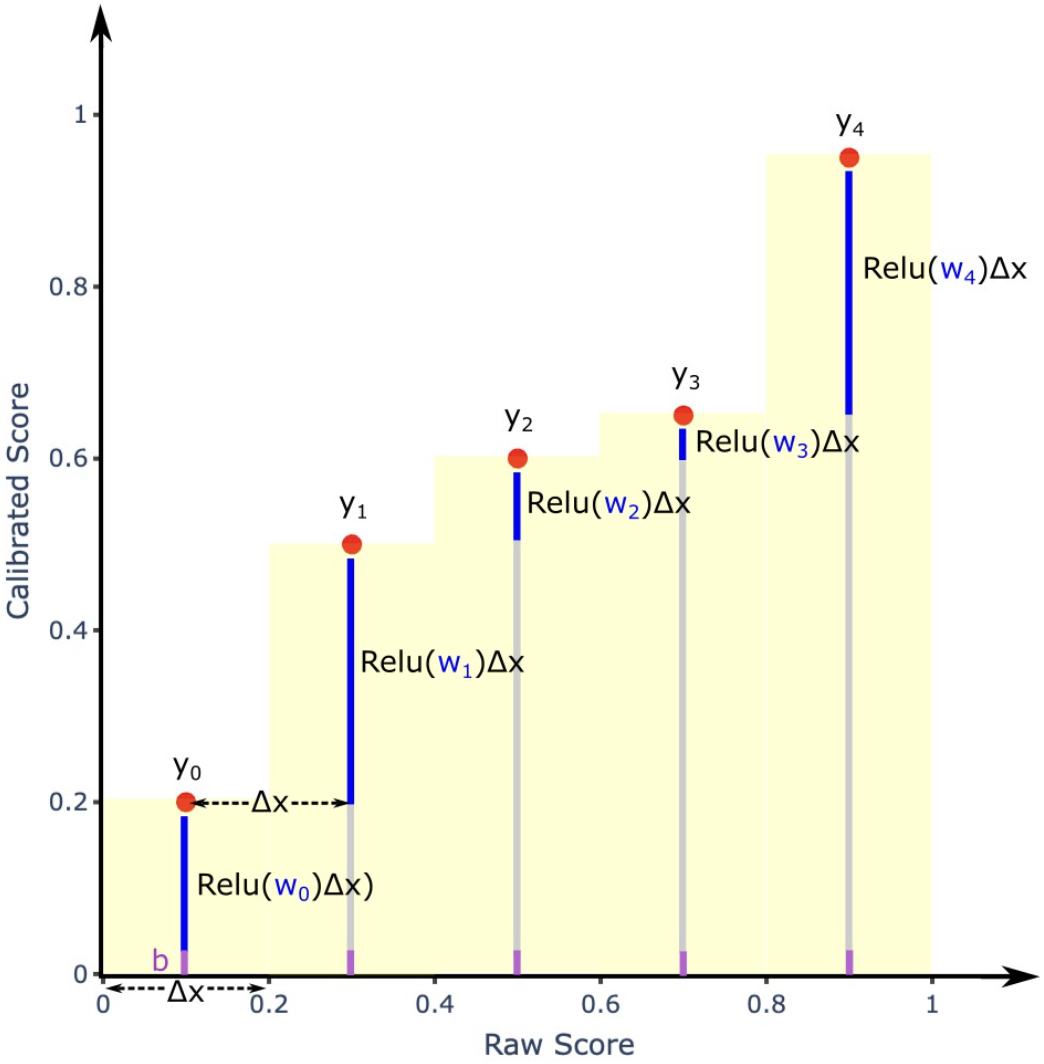}
\caption{Isotonic layer representation}
\label{fig:isotonic_layer}
\end{figure}

To address the issues mentioned above, we developed a customized isotonic regression layer (referred as \textit{isotonic layer}) that can be used as a native neural network layer to be co-trained with a deep neural network model to perform calibration. Similar to the isotonic regression, the isotonic layer follows the piece-wise fitting idea. It bucketizes the predicted values (probabilities must be converted back to logits) by a given interval $v_i$ and assigns a trainable weight $w_i$ for each bucket, which are updated during the training with other network parameters (Figure \ref{fig:isotonic_layer}). The isotonic property is guaranteed by using non-negative weights, which is achieved by using the Relu activation function. To enhance its calibration power with multiple features, the weights can be combined with an embedding representation (a vector whose element is denoted as $e_i$) that derives from all calibration features. 
Finally we obtain
\begin{equation} \label{eq:calibration}
\vspace{-1em}
\begin{aligned}
y_{cali} & = \Sigma_{i=0}^{i=k}Relu(e_i + w_i) \cdot v_i + b, 
v_{i}      =  \begin{cases}  step, & \text{if } i < k \\ y - step \cdot k, & \text{i=k}  \end {cases},  \\ 
k        & = \arg \max_{j}({y-step \cdot j}>0).
\vspace{-1em}
\end{aligned}
\end{equation}

%% file: dense_gating.tex
\subsection{Dense Gating and Large MLP}\label{sec:dense_gating}
Introducing personalized embeddings to global models helps introduce interactions among existing dense features, most of them being multi-dimensional count-based and categorical features. We flattened these multi-dimensional features into a singular dense vector, concatenating it with embeddings before transmitting it to the MLP layers for implicit interactions. A straightforward method to enhance gain was discovered by enlarging the width of each MLP layer, fostering more comprehensive interactions. For Feed, the largest MLP configuration experimented with offline was 4 layers of width 3500 each (refer as "Large MLP", or LMLP). Notably, gains manifest online exclusively when personalized embeddings are in play. However, this enhancement comes at the expense of increased scoring latency due to additional matrix computations. To address this issue, we identified a optimal configuration that maximizes gains within the latency budget.

Later, inspired by Gate Net \cite{DBLP:journals/corr/abs-2007-03519}, we introduced a gating mechanism to hidden layers. This mechanism regulates the flow of information to the next stage within the neural network, enhancing the learning process. We found that the approach was most cost-effective when applied to hidden layers, introducing only negligible extra matrix computation while consistently producing online lift.

Additionally we have explored sparse gated mixture of expert models (sMoE) ~\cite{conf/iclr/ShazeerMMDLHD17}. We report ablation studies in \S\ref{sec:feed_ablation}.

%% file: incremental_training.tex
\subsection{Incremental Training}\label{sec:incremental_training}

Large-scale recommender systems must adapt to rapidly evolving ecosystems, constantly incorporating new content such as Ads, news feed updates, and job postings. To keep pace with these changes, there is a temptation to use the last trained model as a starting point and continue training it with the latest data, a technique known as \textit{warm start}. While this can improve training efficiency, it can also lead to a model that forgets previously learned information, a problem known as catastrophic forgetting\cite{goodfellow2013empirical}. Incremental training, on the other hand, not only uses the previous model for weight initialization but also leverages it to create an informative regularization term.

Denote the current dataset at timestamp $t$ as $\mathcal{D}_t$, the last estimated weight vector as $\mathbf{w}_{t-1}$, the Hessian matrix with regard to $\mathbf{w}_{t-1}$ as $\mathcal{H}_{t-1}$. The total loss up to timestamp $t$ is approximated as
\begin{equation} \label{eq:reg}
\vspace{-0.1em}
\text{loss}_{\mathcal{D}_t}(\mathbf{w}) + \lambda_f/2\times(\mathbf{w}-\mathbf{w}_{t-1})^T\mathcal{H}_{t-1}(\mathbf{w}-\mathbf{w}_{t-1}),
\vspace{-0.1em}
\end{equation}
where $\lambda_f$ is the forgetting factor for adjusting the contribution from the past samples. In practice $\mathcal{H}_{t-1}$ will be a very large matrix. Instead of computing  $\mathcal{H}_{t-1}$, we only use the diagonal elements $\text{diag}(\mathcal{H}_{t-1})$, which significantly reduces the storage and the computational cost. For large deep recommendation models, since the second order derivative computation is expensive, Empirical Fisher Information Matrix (FIM) \cite{pascanu2013revisiting, Kirkpatrick:16} is proposed to approximate the diagonal of the Hessian. 

A typical incremental learning cycle consists of training one initial cold start model and training subsequent incrementally learnt models. To further mitigate catastrophic forgetting and address this issue, we use both the prior model and the initial cold start model to initialize the weights and to calculate the regularization term. In this setting, the total loss presented in \eqref{eq:reg} is:  
\begin{equation} \label{eq:coldweight}
\begin{aligned}
&\text{loss}_{\mathcal{D}_t}(\mathbf{w}) + \lambda_f/2\times[\alpha(\mathbf{w}-\mathbf{w}_{0})^T\mathcal{H}_{0}(\mathbf{w}-\mathbf{w}_{0}) \\
&+ (1-\alpha)(\mathbf{w}-\mathbf{w}_{t-1})^T\mathcal{H}_{t-1}(\mathbf{w}-\mathbf{w}_{t-1})],
\end{aligned}
\end{equation}
where $\mathbf{w}_0$ is the weight of the initial cold start model and $\mathcal{H}_{0}$ is the Hessian with regard to $\mathbf{w}_0$ over the cold start training data. Model weight $\mathbf{w}$ is initialized as $\alpha \mathbf{w_0} + (1-\alpha)\mathbf{w}_{t-1}$. The additional tunable parameter $\alpha \in [0, 1]$ is referred to as \textit{cold weight} in this paper. Positive cold weight continuously introduces the information of the cold start model to incremental learning. When cold weight is $0$, then equation \eqref{eq:coldweight} is the same as \eqref{eq:reg}.

%% file: transact.tex
\subsection{Member History Modeling}\label{sec:trans_act}
To model member interactions with platform content, we adopt an approach similar to \cite{xia2023transact, chen2019behavior}. We create historical interaction sequences for each member, with item embeddings learned during optimization or via a separate model, like \cite{pancha2022pinnerformer}. These item embeddings are concatenated with action embeddings and the embedding of the item currently being scored (early fusion). A Transformer-Encoder \cite{vaswani2017attention} processes this sequence, and the max-pooling token is used as a feature in the ranking model. To enhance information, we also consider the last five sequence steps, flatten and concatenate them as additional input features for the ranking model.
From an ablation study we found that the optimal learning rate for the model with TransAct was similar to the model without TransAct. For the number of encoder layers, going from zero (just pooling) to one layer provides the largest gains, one to two layers smaller gains, and no additional gains beyond three layers. When changing the feedforward dimension as multiples of the embedding size we observe slight additional gains by going from 1/2x to 1x, 2x, and 4x. Similar trends were observed for increasing the sequence length from 25, to 50, to 100. To optimize for latency we used two encoder layers, feedforward dimension as 1/2x the embedding dimension, and sequence length 50. In ablation experiments in \S\ref{sec:feed_ablation} we refer to history modeling as TransAct.

%% file: explore_exploit.tex
\subsection{Explore and Exploit}\label{sec:explore_exploit}
The exploration vs exploitation dilemma is common in recommender systems. A simple utilization of member’s historical feedback data ("exploitation") to maximize immediate performance might hurt long term gain; while boosting new items (“exploration”) could help improve future performance at the cost of short term gain.  To balance them, the traditional methods such as Upper Confidence Bounds (UCB) and Thompson sampling are utilized, however, they can’t be efficiently applied to deep neural network models. To reduce the posterior probability computation cost and maintain certain representational power, we adopted a method similar to the Neural Linear method mentioned in the paper \cite{riquelme2018deep}, namely we performed a Bayesian linear regression on the weights of the last layer of a neural network. Given a predicted value $y_i$ for each input $x_i$ is given by $y_i = W Zx$, where $W$ is the weights of last layer and $Zx$ is the input to the last layer given input $x$. Given  $W$ we apply a Bayesian linear regression to $y$ with respect to $Zx$, and acquire the posterior probability of $W$, which is fed into Thompson Sampling. Unlike the method mentioned in the paper, we don’t independently train a model to learn a representation for the last layer. The posterior probability of W is incrementally updated at the end of each offline training in a given period, thus frequent retrainings would capture new information timely. The technique has been applied to feed and online A/B testing showed relative +0.06\% professionals Daily Active Users.

%% file: lpm_wide.tex
\subsection{Wide Popularity Features} \label{sec:popularity_features}

Our ranking model combines a global model with billions of parameters to capture broad trends and a random effect model to handle variations among individual items, assigning unique values reflecting their popularity among users. Due to our platform's dynamic nature, random effect models receive more frequent training to adapt to shifting trends.

For identifiers with high volatility and short-lived posts, known as Root Object ID, we use a specialized Root-object (RO) model. This model is trained every 8 hours with the latest data to approximate the residuals between the main model's predictions and actual labels. Due to higher coverage of labels we used Likes and Clicks within RO model.
\begin{figure}[ht]
\centering
\includegraphics[width=0.5\textwidth]{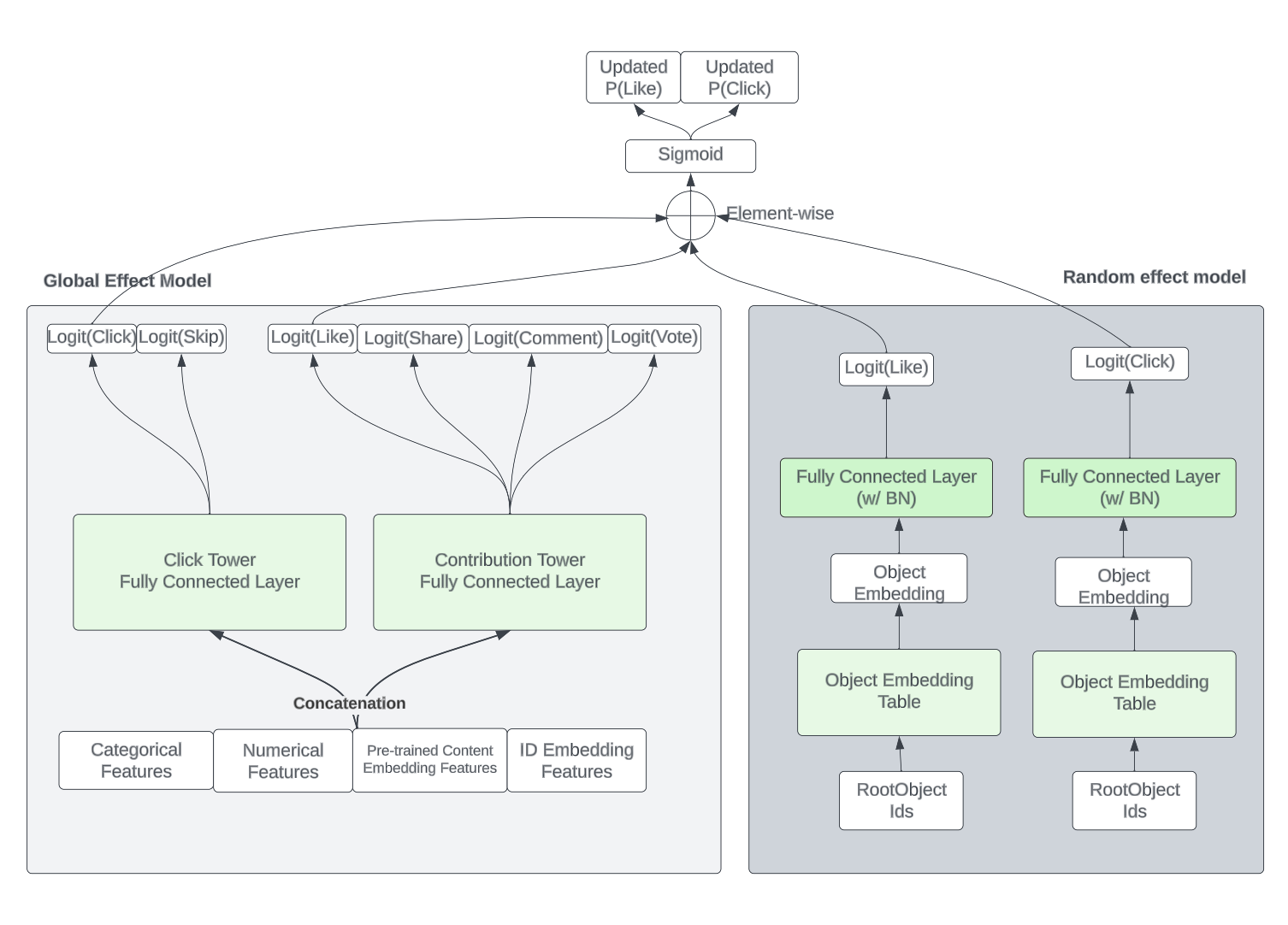}
\caption{RO Wide model on click and like towers.}
\label{fig:ro_wide}
\end{figure}

The final prediction of our model, denoted as \( y_{\text{final}} \), hinges on the summation of logits derived from the global model and the random effect model. It is computed as follows:
\begin{align*}
y_{\text{final}} &= \sigma\left(\text{logit}(y_{\text{global\_effect}}) + \text{logit}(y_{\text{random\_effect}})\right),
\end{align*}
where \( \sigma \) signifies the sigmoid function.

Large embedding tables aid our item ID learning process. We've incorporated an explore/exploit algorithm alongside RO Wide scores, improving the Feed user experience with +0.17\% relative increase in engaged DAU (daily active users).

%% file: multi_task_learning.tex
\subsection{Multi-task Learning}\label{sec:multitask}
Multi-task Learning (MTL) is pivotal for enhancing modern feed ranking systems, particularly in Second Pass Ranking (SPR). MTL enables SPR systems to optimize various ranking criteria simultaneously, including user engagement metrics, content relevance, and personalization.
Our exploration of MTL in SPR has involved various model architectures designed to improve task-specific learning, each with unique features and benefits: (1) Hard Parameter Sharing: involves sharing parameters directly across tasks, serving as a baseline, (2) Grouping Strategy: tasks are grouped based on similarity, such as positive/negative ratio or semantic content. For example, tasks like `Like' and `Contribution' are can be grouped together into a single tower supporting both tasks due to their higher positive rates, while `Comment' and `Share' are grouped separately with lower positive rates.
We also explored common approaches, including MMoE \cite{ma2018modeling} and PLE \cite{tang2020progressive}. In our experiments, the Grouping Strategy showed a modest improvement in metrics with only a slight increase in model parameters (see Table \ref{table:mtl_models}). On the other hand, MMoE and PLE, while offering significant performance boosts, expanded the parameter count by 3x-10x, depending on the expert configuration, posing challenges for large-scale online deployment.
\begin{table}[ht]
\centering
\small 
\begin{tabular}{lccc}
\hline
\textbf{Model} & \textbf{Contributions} \\ \hline
Hard Parameter Sharing & baseline \\
Grouping Strategy & +0.75\% \\
MMoE & +1.19\% \\
PLE & +1.34\% \\ \hline
\end{tabular}
\caption{Performance comparison of MTL models}
\label{table:mtl_models}
\vspace{-3.0em}
\end{table}

%% file: dwell_modeling.tex
\subsection{Dwell Time Modeling}\label{sec:dwell_modeling}

Dwell time, reflecting member content interaction duration, provides valuable insights into member's behavior and preferences. We introduced a `long dwell' signal to detect passive content consumption on the LinkedIn Feed. Implementing this signal effectively, allows the capture of passive but positive engagement. Modeling dwell time presented technical challenges: (1) Noisy dwell time data made direct prediction or logarithmic prediction unsuitable due to high volatility, (2) Static threshold identification for `long dwell' couldn't adapt to evolving user preferences, manual thresholds lacked consistency and flexibility, (3) Fixed thresholds could bias towards content with longer dwell times, conflicting with our goal of promoting engaging posts across all content types on LinkedIn Feed.

To address these challenges, we designed a `long dwell' binary classifier predicting whether there is more time spent on a post than a specific percentile (e.g., 90th percentile). Specific percentiles are determined based on contextual features such as ranking position, content type, and platform, forming clusters for long-dwell threshold setting and enhancing training data. By daily measuring cluster distributions, we capture evolving member consumption patterns and reduce bias and noise in the dwell time signal. The model operates within a Multi-task multi-class framework, resulting in relative improvements of a 0.8\% in overall time spent, a 1\% boost in time spent per post, and a 0.2\% increase in member sessions.

%% file: qr_and_murmur.tex
\subsection{Model Dictionary Compression}\label{sec:qr_and_murmur}
The traditional approach to mapping high-dimensional sparse categorical features to an embedding space involves two steps. First, it converts string-based ID features to integers using a static hashtable. Next, it utilizes a memory-efficient Minimal Perfect Hashing Function (MPHF) \cite{Limasset2017MPH} to reduce in-memory size. These integer IDs serve as indices for accessing rows in the embedding matrix, with cardinality matching that of the static hashtable or unique IDs in the training data, capped at a maximum limit. The static hashtable contributes for about 30\% of memory usage, which can become inefficient as vocabulary space grow and the vocabulary-to-model size ratio increases. Continuous training further complicates matters, as it demands incremental vocabulary updates to accommodate new data.

QR hashing \cite{DBLP:journals/corr/abs-1909-02107} offers a solution by decomposing large matrices into smaller ones using quotient and remainder techniques while preserving embedding uniqueness across IDs. For instance, a vocabulary of 4 billion with a 1000x compression ratio in a QR strategy results in two tiny embedding matrices of approximately 4 million rows in sum — roughly 4 million from the quotient matrix and around 1000 from the remainder matrix. This approach has demonstrated comparable performance in offline and online metrics in Feed/Ads. We found that sum aggregation worked the best, while multiplication aggregation suffered from convergence issues due to numerical precision, when embeddings are initialized close to 0. QR hashing's compatibility with extensive vocabulary opens doors to employing a collision-resistant hashing function like MurmurHash, potentially eliminating vocabulary maintenance. It also generates embedding vectors for every training item ID, resolving the Out-of-Vocabulary (OOV) problem and can potentially capture more diverse signals from the data. Refer Figure \ref{fig:LMP_hashing} in Appendix for illustration on the technique.
\vspace{-1.0em}

%% file: model_quantization.tex
\subsection{Embedding Table Quantization}\label{sec:model_quantization}
Embedding tables, often exceeding 90\% of a large-scale deep ranking model's size, pose challenges with increasing feature, entity, and embedding dimension sizes. These components can reach trillions of parameters, causing storage and inference bottlenecks due to high memory usage~\cite{guan2019post} and intensive lookup operations. To tackle this, we explore embedding table quantization, a model dictionary compression method that reduces embedding precision and overall model size. For example, using an embedding table of 10 million rows by 128 with fp32 elements, 8-bit row-wise min-max quantization \cite{Shen_Dong_Ye_Ma_Yao_Gholami_Mahoney_Keutzer_2020} can reduce the table size by over 70\%. Research has shown that 8-bit post-training quantization maintains performance and inference speed without extra training costs or calibration data requirements~\cite{guan2019post}, unlike training-aware quantization. To ensure quick model delivery, engineer flexibility, and smooth model development and deployment, we opt for post-training quantization, specifically employing middle-max row-wise embedding-table quantization.
Unlike min-max row-wise quantization which saves the minimum value and the quantization bin-scale value of each embedding row, middle-max quantization saves the middle values of each row defined by {\small $\mathbf{X}_{i, :}^{middle} = \frac{\mathbf{X}_{i, :}^{max} * 2^{bits-1} + \mathbf{X}_{i, :}^{min} * (2^{bits-1} - 1)}{2^{bits} - 1}$}, where $\mathbf{X}_{i, :}^{min}$ and $\mathbf{X}_{i, :}^{max}$ indicate the minimum and maximum value of the $i$-th row of an embedding table $\mathbf{X}$. The quantization and dequantization steps are described as: 
    $\mathbf{X}_{i, :}^{int} = round(\frac{\mathbf{X}_{i, :}-\mathbf{X}_{i, :}^{middle}}{\mathbf{X}_{i, :}^{scale}})$ and $\mathbf{X}_{i, :}^{dequant} = \mathbf{X}_{i, :}^{middle} + \mathbf{X}_{i, :}^{int} * \mathbf{X}_{i, :}^{scale} $, 
    where $\mathbf{X}_{i, :}^{scale}= \frac{\mathbf{X}_{i, :}^{max} - \mathbf{X}_{i, :}^{min} }{ 2^{bits} - 1 }$.

We choose middle-max quantization for two reasons: (1) Embedding values typically follow a normal distribution, with more values concentrated in the middle of the quantization range. Preserving these middle values reduces quantization errors for high-density values, potentially enhancing generalization performance. (2) The range of $\mathbf{X}_{i, :}^{int}$ values falls within $[-128, 127]$, making integer casting operations from float to int8 reversible and avoiding 2's complement conversion issues, i.e., cast(cast(x, int8), int32) may not be equal to x due to the 2's complement conversion if $x\in [0, 255]$. Experimental results show that 8-bit quantization generally achieves performance parity with full precision, maintaining reasonable serving latency even in CPU serving environments with native TF operations. In Ads CTR prediction, we observed a +0.9\% CTR relative improvement in online testing, which we attribute to quantization smoothing decision boundaries, improving generalization on unseen data, and enhancing robustness against outliers and adversaries.

%% file: training_scalability.tex
\section{Training scalability}\label{sec:training_scalability}

During development of large ranking models we optimized training time via set of techniques including 4D Model Parallelism, Avro Tensor Dataset Loader, offloading last-mile transformation to async stage and prefetching data to GPU with significant improvements to training speed (see Table \ref{table:Training_optimizations_table}). Below we provide descriptions on why and how we developed it.

\subsection{4D Model Parallelism}
We utilized Horovod to scale out synchronous training with multiple GPUs. During benchmarking, we have observed performance bottlenecks during gradient synchronization of the large embedding tables. We implemented 4D model parallelism in TensorFlow (TF) to distribute the embedding table into different processes. Each worker process will have one specific part of the embedding table shared among all the workers. We were able to reduce the gradient synchronization time by exchanging input features via all-to-all (to share the features related to the embedding lookup to specific workers), which has a lower communication cost compared to exchanging gradients for large embedding tables. From our benchmarks, model parallelism reduced training time from 70 hours to 20 hours.

\subsection{Avro Tensor Dataset Loader}
We also implemented and open sourced a \href{https://github.com/tensorflow/io/blob/075be7222dfd234c902aeb31e2e0a44a8db49c00/AVRO_TENSOR_DATASET.md}{TF Avro reader} that is up to 160x faster than the existing Avro dataset reader according to our benchmarks. Our major optimizations include removing unnecessary type checks, fusing I/O operations (parsing, batching, shuffling), and thread auto-balancing and tuning. With our dataset loader, we were able to resolve the I/O bottlenecks for training job, which is common for large ranking model training. The e2e training time was reduced by ~50\% according to our benchmark results (Table \ref{table:Training_optimizations_table}).

\subsection{Offload Last-mile Transformation to Asynchronous Data Pipeline}
We observed some last-mile in-model transformation that happens inside the training loop (ex. filling empty rows, conversion to Dense, etc.). Instead of running the transformation + training synchronously in the training loop, we moved the non-training related transformation to a transformation model, and the data transformation is happening in the background I/O threads that is happening asynchronously with the training step. After the training is finished, we stitched the two model together into the final model for serving. The e2e training time was reduced by ~20\% according to our benchmark results (Table \ref{table:Training_optimizations_table}).

\subsection{Prefetch Dataset to GPU}
During the training profiling, we saw CPU -> GPU memory copy happens during the beginning of training step. The memory copy overhead became significant once we increased the batch size to larger values (taking up to 15\% of the training time). We utilized customized TF dataset pipeline and Keras Input Layer to prefetch the dataset to GPU in parallel before the next training step begins.

\begin{table}[ht]
\centering
\small 
\begin{tabular}{lccc}
\hline
\textbf{Optimization Applied} & \textbf{e2e Training Time Reduction} \\ \hline
4D Model Parallelism & 71\% \\
Avro Tensor Dataset Loader & 50\%\\ 
Offload last-mile transformation & 20\%\\ 
Prefetch dataset to GPU & 15\% \\ 
\hline
\end{tabular}
\caption{Training performance relative improvements}
\label{table:Training_optimizations_table}
\vspace{-2.5em}
\end{table}

%% file: Experiments.tex
\section{Experiments} \label{sec:experiments}
We conduct offline ablation experiments and A/B tests across various surfaces, including Feed Ranking, Ads CTR prediction, and Job recommendations. In Feed Ranking, we rely on offline replay metrics, which have shown a correlation with production online A/B test results. Meanwhile, for Ads CTR and Job recommendations, we find that offline AUC measurement aligns well with online experiment outcomes.
\subsection{Incremental Learning}
We tested incremental training on both Feed ranking models and Ads CTR models. The experiment configuration is set in Table \ref{table:it_experiments}. We start with a cold start model, followed by a number of incremental training iterations (6 for Feed ranking models and 4 for Ads CTR models). For each incrementally trained model, we evaluate on a fixed test dataset and average the metrics. The baseline is the evaluation metric on the same fixed test dataset using the cold start model. 
\begin{table}[h]
\small
\begin{tabular}{ccc}
\hline
                 Experiments      & Feed Ranking  & Ads CTR \\ \hline
Cold Start Data Range  & 21 days                  & 14 days             \\ 
Incremental Data Range & 1 day                    & 0.5 day             \\
Incremental Iterations & 6                        & 4   \\ \hline               
\end{tabular}
\caption{Incremental Experiments Settings}
\label{table:it_experiments}
\vspace{-2.5em}
\end{table}

Table \ref{table:deep_feeds} and \ref{table:deep_ads} summarize the metrics improvements and training time improvements for both Feed ranking models and Ads CTR models, after tuning the cold weight and $\lambda$. For both models, incremental training boosted metrics with significant training time reduction. Contributions measurement for Feed is explained in \S\ref{sec:feed_ablation}.

\begin{table}[h]
\small
\begin{tabular}{ccc}
\hline
         & Contributions & Training Time \\ \hline
Cold Start & -  & - \\
Incremental Training & +1.02\% & -96\% \\ \hline
\end{tabular}
\caption{Feed ranking model results summary}
\label{table:deep_feeds}
\vspace{-2.5em}
\end{table}

\begin{table}[h]
\small
\begin{tabular}{ccc}
\hline
            & Test AUC & Training Time \\ \hline
Cold Start  & -  & - \\ 
Incremental Training & +0.18\% & -96\% \\ \hline
\end{tabular}
\caption{Ads CTR model results summary}
\label{table:deep_ads}
\vspace{-2.5em}
\end{table}

\input{feed_ranking_metrics.tex}
\input{job_recommendations.tex}
\input{ads_ctr.tex}

%% file: feed_ranking_metrics.tex
\subsection{Feed Ranking}\label{sec:feed_ablation}
To assess and compare Feed ranking models offline, we employ a "replay" metric that estimates the model's online contribution rate (e.g., likes, comments, re-posts). For evaluation, we rank a small portion of LinkedIn Feed sessions using a pseudo-random ranking model, which uses the current production model to rank all items but randomizes the order of the top N items uniformly. After training a new experimental model, we rank the same sessions offline with it. When a matched impression appears at the top position ("matched imp @ 1," meaning both models ranked the same item at Feed position 1) and the member served the randomized model makes a contribution to that item, we assign a contribution reward to the experimental model:
$\text{contribution rate} = \frac{\text{\# of matched imps @ 1 with contribution}}{\text{\# of matched imps @ 1}}$

This methodology allows unbiased offline comparison of experimental models \cite{Li:2011:UOE:3045725.3045727}. We use offline replay to assess Feed Ranking models, referred to as 'contribution' throughout the paper (Table \ref{table:feed_ablation_models}). The table illustrates the impact of various production modeling techniques on offline replay metrics, including Isotonic calibration layer, low-rank DCNv2, Residual DCN, Dense Gating, Large MLP layer, Sparse Features, MTL enhancements, TransAct, and Sparsely Gated MMoE. These techniques, listed in Table \ref{table:feed_ablation_models}, are presented in chronological order of development, highlighting incremental improvements. We have deployed these techniques to production, and through online A/B testing, we observed a 0.5\% relative increase in the number of member sessions visiting LinkedIn.
\begin{table}[]
\begin{tabular}{@{}lccc@{}}
\toprule
\textbf{Model}               & \textbf{Contributions} & \textbf{\begin{tabular}[c]{@{}c@{}}Latency\\ (p90)\end{tabular}} & \textbf{\begin{tabular}[c]{@{}c@{}}CPU\\ Usage\\ (p95)\end{tabular}} \\ \midrule
Baseline                     & --                     & --                                                               & --                                                                   \\
+ 30-dim ID embeddings       & $+1.89\%$              & --                                                               & $+20\%$                                                              \\
+ Isotonic calibration layer & $+1.08\%$              & --                                                               & --                                                                   \\
+ Large MLP                  & $+1.23\%$              & --                                                               & $+17\%$                                                              \\
+ Dense Gating               & +1.00\%                & --                                                               & --                                                                   \\
+ Multi-task Grouping        & $+0.75\%$              & --                                                               & --                                                                   \\
+ Low-rank DCNv2             & $+1.26\%$              & --                                                               & $+13\%$                                                              \\
+ TransAct                   & $+1.66\%$              & $+52\%$                                                          & $+44\%$                                                              \\
+ Residual DCN               & $+2.15\%$              & --                                                               & $+17\%$                                                              \\
+ LDCNv2+LMLP+TransAct       & $+3.45\%$              & N/A                                                              & N/A                                                                  \\
+ RDCN+LMLP+TransAct         & $+3.62\%$              & N/A                                                              & N/A                                                                  \\
+ Sparsly Gated MMoE         & +4.14\%                & N/A                                                              & N/A                                                                  \\ \bottomrule
\end{tabular}
\caption{Feed ranking ablation study results. Given are percentage increases for contributions, latency, and CPU usage. A dash "--" indicates neutrality.}
\end{table}

%% file: job_recommendations.tex
\subsection{Jobs Recommendations}\label{sec:job_recommendations}

In Job Search (JS) and Jobs You Might Be Interested In (JYMBII) ranking models, 40 categorical features are embedded through 5 shared embedding matrices for title, skill, company, industry, and seniority. The model predicts probability of P(job application) and P(job click). We adopted embedding dictionary compression described in \S\ref{sec:qr_and_murmur} with 5x reduction of number of model parameters, and the evaluation does not show any performance loss compared to using vanilla id embedding lookup table. We also did not observe improvement by using Dense Gating (\S\ref{sec:dense_gating}) in JYMBII and JS with extensive tuning of models. These entity id embeddings are shared by Job Search and JYMBII Recommendation, and then a task-specific 2-layer DCN is added on top to explicitly capture the feature interactions. Overall we observe significant offline AUC lift of +1.63\% for Job Search and 2.10\% for JYMBII. For reproducibility purposes we provide model architecture and ablation study of different components of JYMBII and Job Search model in \S\ref{sec:appendix:job_rec}.

The ranking models with higher AUC shown above also transferred to significant metrics lift in 
the online A/B testing, leading to relative 1.76\% improvement in Qualified Applications across Job Search and JYMBII. Percent Chargeable Views is the fraction of clicks among all clicks on promoted jobs. Qualified Application is the total count of all qualified job applications. 
\begin{table}[ht]
\centering
\small 
\begin{tabular}{lcc}
\hline
\textbf{Online Metrics} & Job Search & JYMBII \\ \hline
Percent Chargeable Views & $+1.70\%$   & $+4.16\%$ \\ 
Qualified Application  & $+0.89\%$   & $+0.87\%$ \\ \hline
\end{tabular}
\caption{Online experiment relative metrics improvements of JS and JYMBII ranking}
\label{table:jobs_recommendations_online_table}
\vspace{-2.5em}
\end{table}

%% file: ads_ctr.tex
\subsection{Ads CTR}
Our baseline model is a multilayer perceptron model that derived from its predecessor GDMix model \cite{kdd2022modeling} with proper hyper-parameter tuning. Features fall into five categories: contextual, advertisement, member, advertiser, ad-member interaction. Baseline model doesn’t have Id features. 
In the Table \ref{table:deep_ads} we show relative improvements of each of the techniques including ID embeddings, Quantization, Low-rank DCNv2, TransAct and Isotonic calibration layer. Techniques mentioned in the table are ordered in timeline of development. 
We have deployed techniques to production and observed 4.3\% CTR relative improvement in online A/B tests. 

\begin{table}[ht]
\centering
\small 
\begin{tabular}{lcc}
\hline
\textbf{Model} & AUC \\ \hline
Baseline & - \\
ID embeddings (IDs)               & +1.27\% \\
IDs+Quantization 8-bit            & +1.28\% \\
IDs+DCNv2                         & +1.45\% \\
IDs+low-rank DCNv2                & +1.37\% \\
IDs+isotonic layer                & +1.39\% \\
                                  & (O/E ratio +1.84\%)\\
IDs+low-rank DCNv2+isotonic layer & +1.47\% \\
IDs + TransAct                    & +2.20\% \\

\hline
\end{tabular}
\caption{Ablation study of different Ads CTR model architecture variants on the test AUC.}
\label{table:ads_ablation_models}
\vspace{-2.0em}
\end{table}

%% file: deployment_lessons.tex
\section{Deployment Lessons}
Over the time of development we learnt many deployment lessons. Here we present couple of interesting examples.
\subsection{Scaling up Feed Training Data Generation}
At the core of the Feed training data generation is a join between post labels and features. The labels dataset consists of impressed posts from all sessions. The features dataset exists on a session level. Here, each row contains session-level features and all served posts with their post-level features. To combine these, we explode the features dataset to be on a post-level and join with the labels dataset. 
However, as Feed scaled up from using 13\% of sessions for training to using 100\% of sessions, this join caused long delay. To optimize the pipeline we made two key changes that reduced the runtime by 80\% and stabilized the job.
Firstly, we recognized that not all served posts are impressed. This means the join with the labels dataset drastically reduces the number of rows. Furthermore, exploding the features dataset repeats session-level features for every post. We therefore changed the pipeline to explode only the post features and keys, join with the labels, and add the session-level features in a second join. Despite this resulting in two joins, each join was now smaller and resulted in an overall shuffle write size reduction of 60\%. Secondly, we tuned the Spark compression, which resulted in an additional 25\% shuffle write size reduction. These changes allowed us to move forward with 100\% of sessions for training.
\vspace{-0.1em}
\subsection{Model Convergence}
Adding DCNv2 came with challenges for model training. During initial training experiments with DCNv2 we observed a large number of runs diverging. To improve model training stability we increased learning rate warm-up from 5\% to 50\% of training steps. This resolved the instability issues and also significantly boosted the offline relevance gains brought about by adding DCNv2. We also applied batch normalization to the numeric input features as suggested in \cite{xia2023transact}. Finally, we found that at our number of training steps we were under-fitting. This became clear when we observed that increasing the training steps significantly improved offline relevance metrics. However, increasing the number of training steps was not an option for production due to the decrease in experimentation velocity. As a solution, we found that given the increased warm-up steps, our training was stable enough for higher learning rates. Increasing the learning rate three-fold allowed us to almost completely bridge any relevance metric gaps we found compared to longer training.

We found that optimization needs varied across different models. While Adam was generally effective, models with numerous sparse features required AdaGrad, which significantly impacted their performance. Furthermore, we employed strategies like learning rate warm-up and gradient clipping, especially beneficial for larger batch sizes, to enhance model generalization. We consistently implemented learning rate warm-up for larger batches, increasing the learning rate over a doubled fraction of steps whenever batch size doubled, but not exceeding 60\% of the total training steps. By doing so, we improved generalization across various settings and narrowed the gap in generalization at larger batch sizes.

%% file: conclusion.tex
In this paper, we introduced the {\systemname} framework, encapsulating our experience in developing state-of-the-art models. We discussed various modeling architectures and their combination to create a high-performance model for delivering relevant user recommendations. The insights shared in this paper can benefit practitioners across the industry. {\systemname} has been deployed in multiple domain applications at LinkedIn, resulting in significant production impact.

%% file: reproducability.tex

\appendix
\section{INFORMATION FOR REPRODUCIBILITY}\label{sec:reproducability}

\subsection{Feed Ranking Sparse ID features}\label{sec:sparse_features}
The sparse id Feed ranking embedding features consist of (1) Viewer Historical Actor Ids, which were frequently interacted in the past by the viewer, analogous to Viewer-Actor Affinity as in \cite{eAffinity_paper}, (2) Actor Id, who is the creator of the post, (3) Actor Historical Actor Ids, which are creators who frequently interacted in the past by the creator of the post, (4) Viewer Hashtag Ids, which were frequently interacted in the past by the viewer, (5) Actor Hashtag Ids, which were frequently interacted in the past by the actor of the post and (6) Post Hashtag Ids (e.g. \#machinelearning).

We used unlimited dictionary sparse ID features explained in \S\ref{sec:qr_and_murmur}. We empirically found 30 dimensions to be optimal for the Id embeddings. The sparse id embedding features mentioned above are concatenated with all other dense features and then passed through a multi-layer perception (MLP) consisting of 4 connected layers, each with output dimension of 100.

\subsection{Vocabulary Compression for Serving Large Models}
The IDs in large personalizing models are often strings and sparse numerical values. If we want to map the unique sparse IDs to embedding index without any collision,  then a lookup table is needed which is typically implemented as a hash table (e.g. std::unordered\_map in TF). These hash tables grow into several GBs and often take up even more memory than the model parameters. To resolve the serving memory issue, we implemented minimal perfect hashing function (MPHF) \cite{Limasset2017MPH} in TF Custom Ops, which reduced the memory usage of vocab lookup by 100x. However, we faced a 3x slowdown in training time as the hashing was performed on the fly as part of training. We observed that the maximum value of our IDs could be represented using int32. To compress the vocabulary without degrading the training time, we first hashed the string id into int32 using \cite{commons_codec}, and then used the map implementation provided by \cite{vigna_fastutil} to store the vocabulary. We used a Spark job to perform the hashing and thus were able to avoid training time degradation. The hashing from string to int32 provided us with 93\% heap size reduction. We didn't observe significant degradation in engagement metrics because of hashing.

The subsequent effort mentioned in section \S\ref{sec:qr_and_murmur} successfully eliminated the static hash table from the model artifact by employing collision-resistant hashing and QR hashing techniques. This removal was achieved without any performance drop, considering both runtime and relevance perspectives.

\subsection{External Serving of ID Embeddings vs In-memory Serving}
\begin{figure}[h]
\centering
\includegraphics[width=0.5\textwidth]{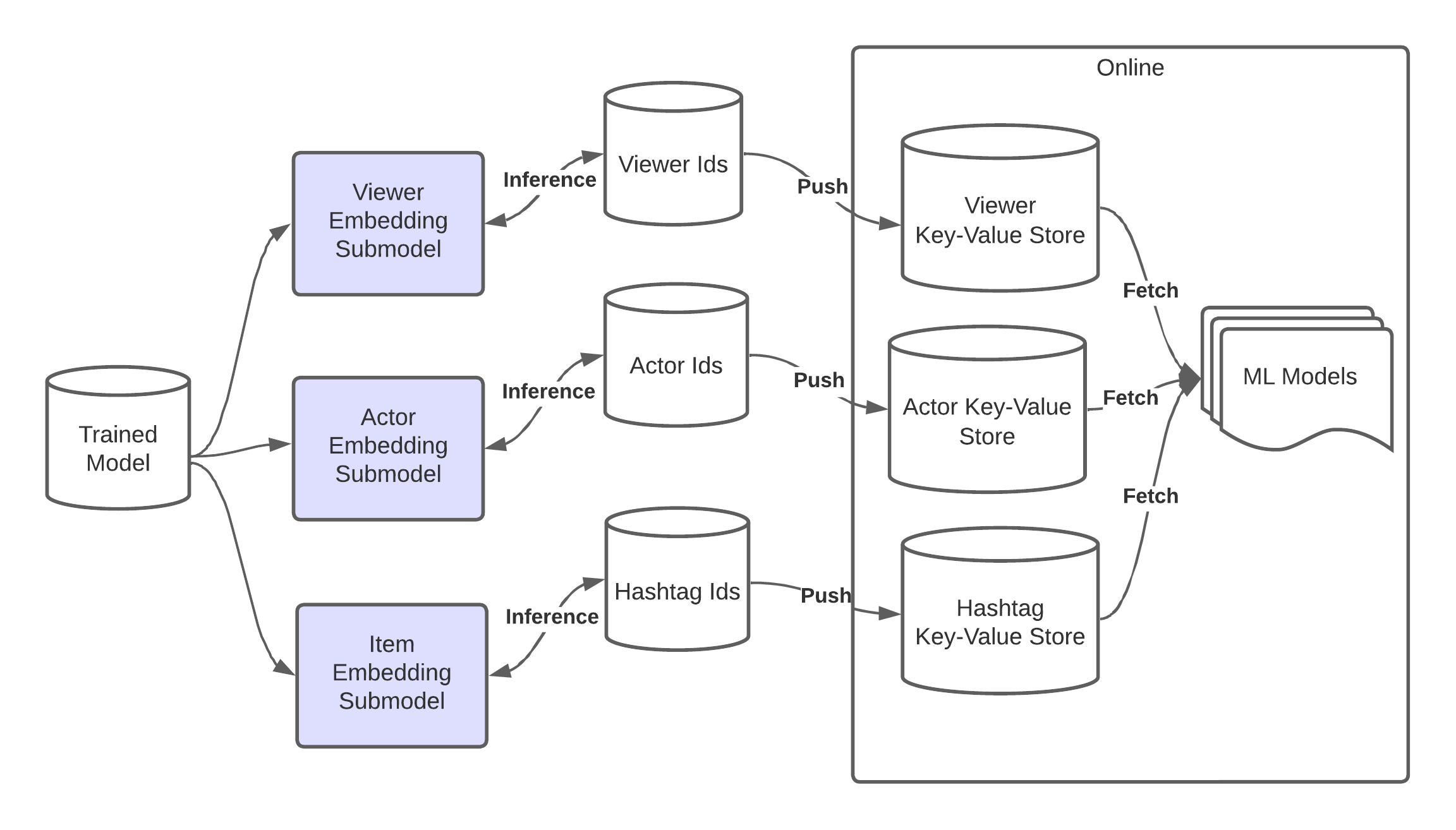}
\caption{External serving strategy \textit{\small{}}}
\label{fig:external_serving}
\vspace{-1.0em}
\end{figure}

One of the challenges was constrained memory on serving hosts, hindering the deployment of multiple models. To expedite the delivery we initially adopted external serving of model parameters in a key-value store (see Figure \ref{fig:external_serving}), partitioning model graphs and pre-computing embeddings for online retrieval. We faced issues with (1) iteration flexibility for ML engineers, who depended on the consumption of ID embeddings, and (2) staleness of pre-computed features pushed daily to the online store. To handle billion-parameter models concurrently from memory, we upgraded hardware and optimized memory consumption by garbage collection tuning, and crafting data representations for model parameters through quantization and ID vocabulary transformation optimized memory usage. As we transitioned to in-memory serving, it yielded enhanced engagement metrics and empowered modelers with reduced operational costs.

\subsection{4D Model Parallelism}
\begin{figure}
 \centering
 \includegraphics[height=5cm]{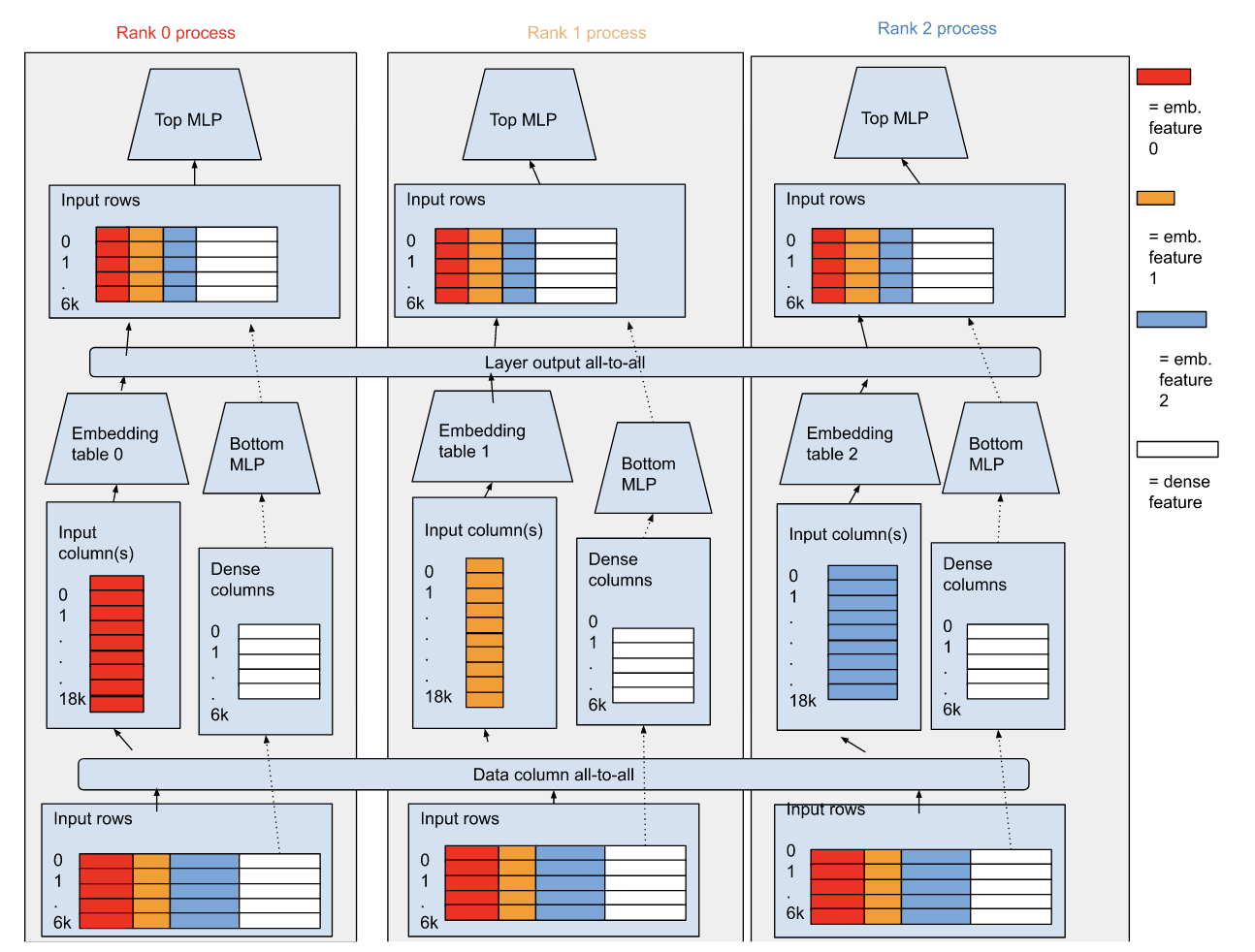}
 \caption{Model parallelism for large embedding tables}
 \label{fig:model_parallelism}
 \vspace{-1.5em}
\end{figure}

\begin{figure}[t]
\centering
\includegraphics[width=0.5\textwidth]{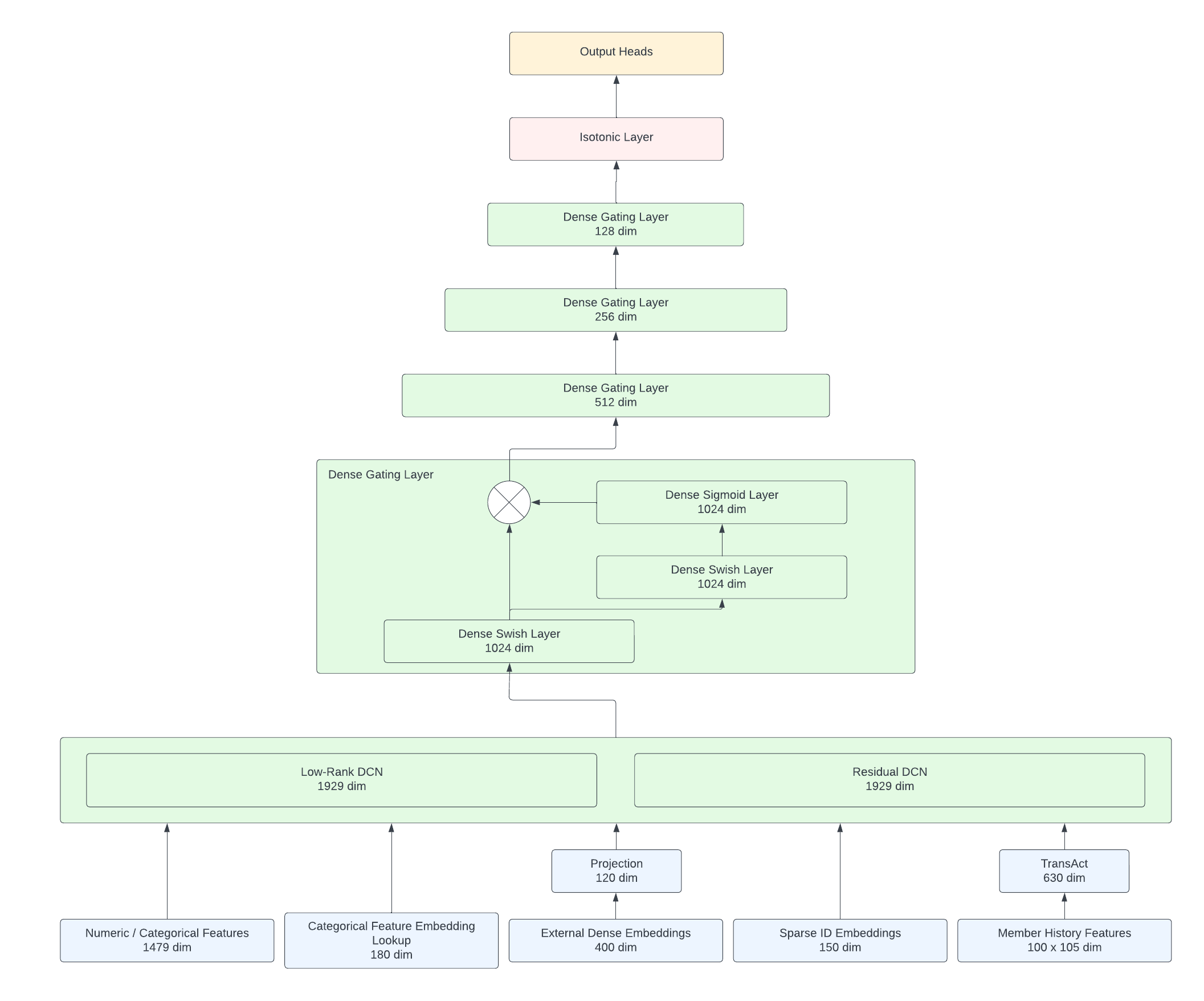}
\caption{Feed ranking model architecture}
\label{fig:feed_model_architecture_details}
\end{figure}

Figure \ref{fig:model_parallelism} shows an example for three embedding tables. Each embedding table is placed on a GPU, and each GPU's input batch is all-to-all'ed so that every GPU receives the input columns belonging to its embedding table. Each GPU does its local embedding lookup, and the lookups are all-to-all'ed to return the output to the GPU that the input column came from. Other layers with fewer parameters (such as MLP layers) are still processed in a data parallel way since exchanging gradients for these layers is not costly. From our benchmarks, model parallelism reduced training time from 70 hours to 20 hours.

\subsection{Experimentation on Sequence Length for User History Models}

Here we present study on how history length influences the impact of the Feed ranking model in Table \ref{table:feed_ablation_models_length}. We observe increasing trend of engagement increase as we use longer history of user engagement over sequence architecture described in \S\ref{sec:trans_act}.
\begin{table}[htbp]
\centering
\small 
\begin{tabular}{lc}
\hline
\textbf{Model} & Contributions \\ \hline
Baseline                                       & -  \\
+ Member history length 25                               & $+1.31\%$  \\
+ Member history length 50                    & $+1.57\%$ \\
+ Member history length 100                      & $+1.66\%$  \\ \hline
\end{tabular}
\caption{Offline relevance metrics for the feed from the addition of member history modeling with different sequence lengths.}
\label{table:feed_ablation_models_length}
\vspace{-2.0em}
\end{table}

\subsection{Feed Ranking Model Architecture}
On the Figure \ref{fig:feed_model_architecture_details} we present Feed Model architecture diagram to provide a flow of the model, and how different parts of the model connected to each other. We found that placement of different modules changes the impact of the techniques significantly.

\subsection{Vocabulary Compression}
\begin{figure}[h]
    \centering
    \includegraphics[width=0.5\textwidth]{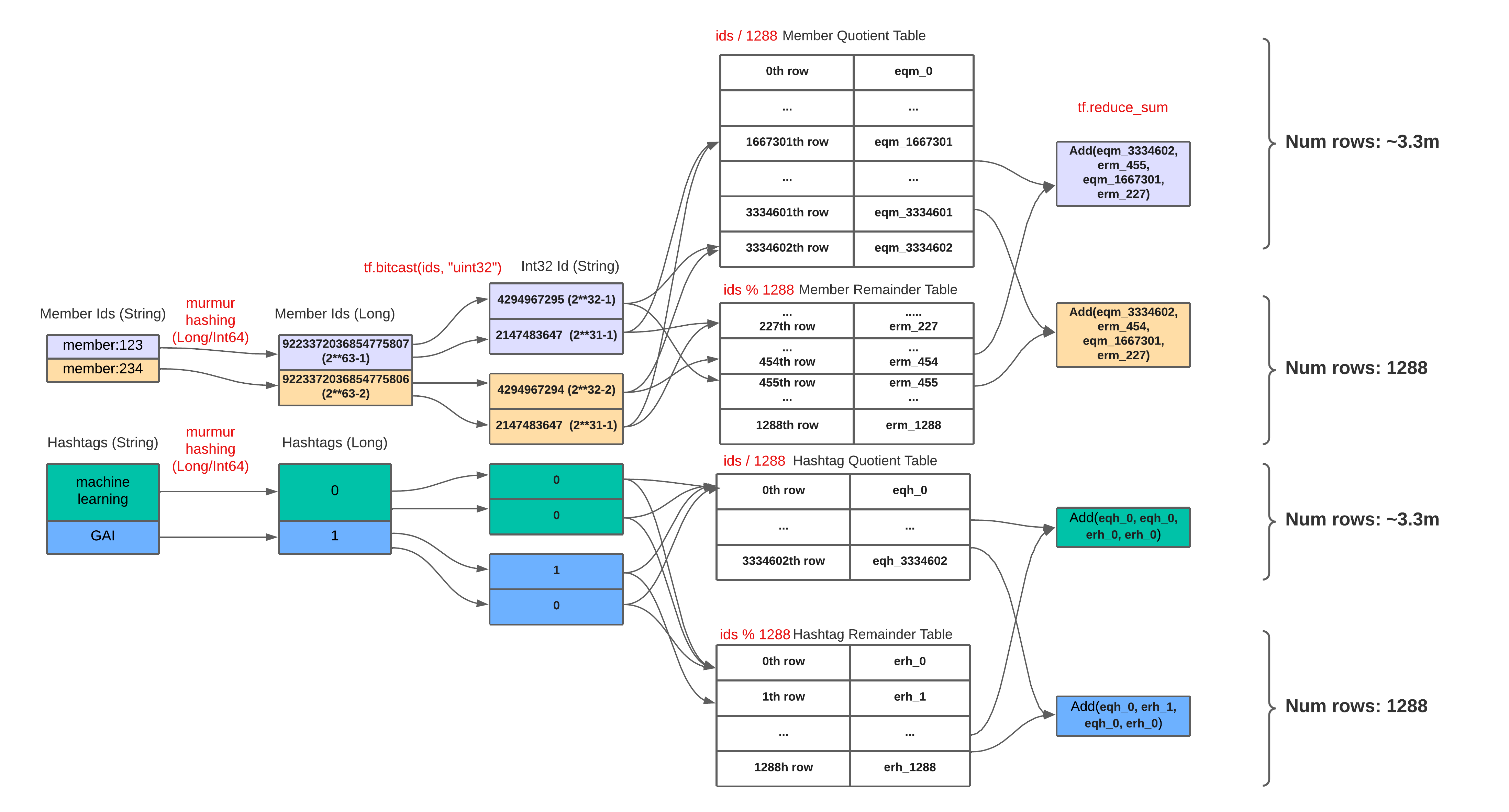}
    \caption{Example of non static vocab hashing paradigm}
    \label{fig:LMP_hashing}
    \end{figure}

On the Figure \ref{fig:LMP_hashing} we present example diagram of non static vocabulary compression using QR and Murmur hashing. A member ID \(A\) in string format like "member:1234," will be mapped with a collision-resistant stateless hashing method (e.g., Murmur hashing) to a space of \texttt{int64} or \texttt{int32}. The larger space will result in a lower collision rate. In our case, we use \texttt{int64}, and then we use \texttt{bitcast} to convert this \texttt{int64} to two numbers in \texttt{int32} space (ranging from 0 to $2^{32}-1$), \(B\) and \(C\) which will look from independent sets of QR tables.

\begin{table}[ht]
\centering
\small 
\begin{tabular}{lcc}
\hline
\textbf{Model} & AUC \\ \hline
Baseline & - \\
IDs + Wide\&Deep \cite{wideDeep2016}             & +0.37\% \\
IDs + Wide\&Deep + Dense Gating (\S\ref{sec:dense_gating})   & +0.33\% \\
IDs + DeepFM \cite{Guo2017DeepFMAF}           & +0.39\% \\
IDs + FinalMLP \cite{Mao2023FinalMLPAE}                        & +2.17\% \\
IDs + DCNv2 \cite{Wang2020DCNVI}                & +2.23\% \\
IDs + DCNv2 + QR hashing (\S\ref{sec:qr_and_murmur})                & +2.23\% \\
\hline
\end{tabular}
\caption{Ablation study of different jobs recommendation model architecture variants on the JYMBII test AUC}
\label{table:jobs_recommendation_ablation_models_}
\vspace{-2.0em}
\end{table}

\subsection{Jobs Recommendations Ranking Model Architecture} \label{sec:appendix:job_rec}
\begin{figure}
\centering
\includegraphics[width=0.5\textwidth]{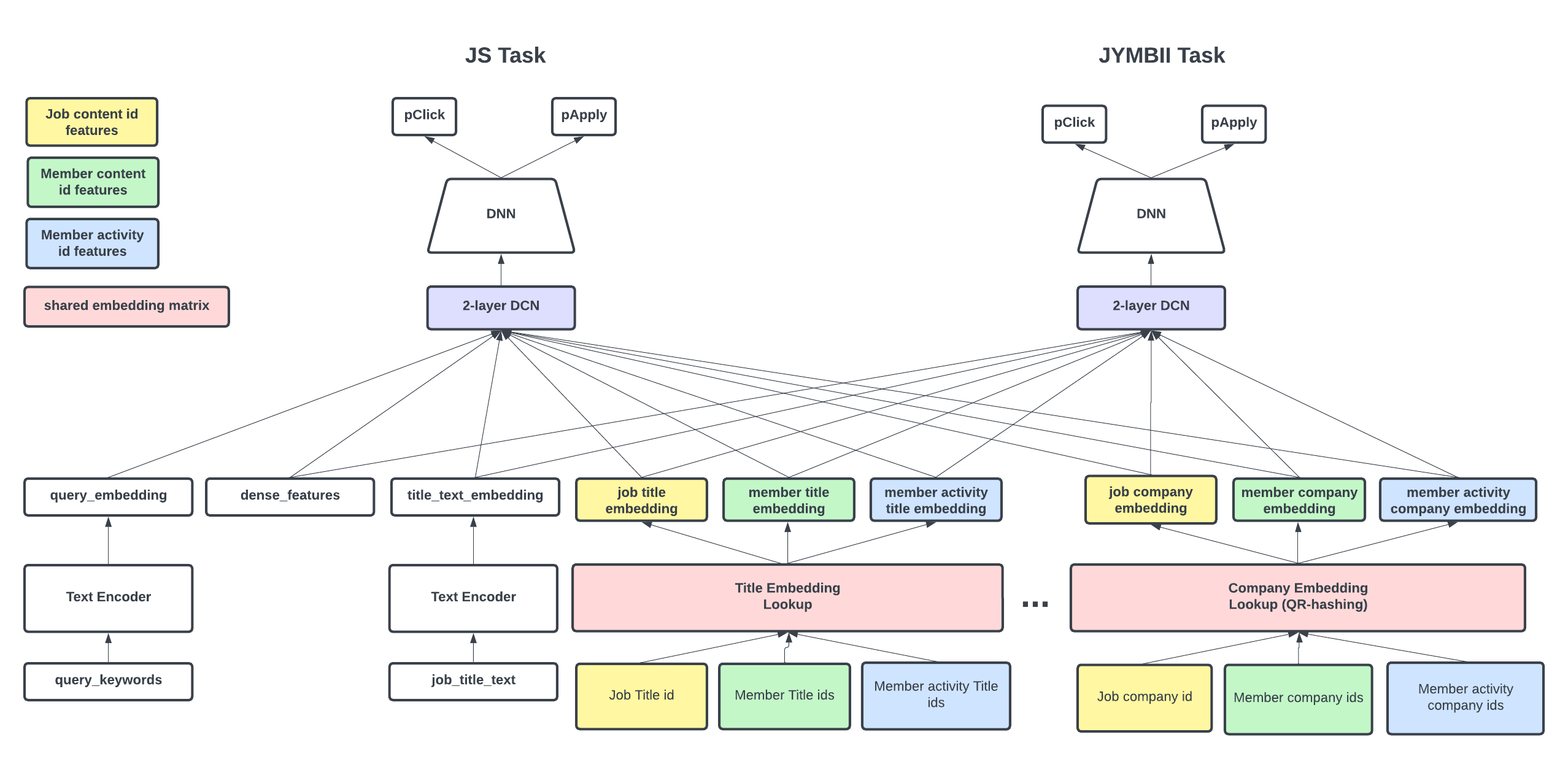}
\caption{Jobs recommendation ranking model architecture}
\label{fig:job_recommendataion_model_architecture}
\end{figure}

As shown in the Figure \ref{fig:job_recommendataion_model_architecture}, the jobs recommendation ranking model employs a multi-tasks training framework that unifies Job Search (JS) and Jobs You Might be Interested In (JYMBII) tasks in a single model. The id embedding matrices are added into the bottom layer to be shared by the two tasks, followed by a task-specific 2-layer DCNv2 to learn feature interactions. We conducted various experiments to apply different architectures of feature interactions, and the 2-layer DCN performs best among all. The results on the JYMBII task are demonstrated in the Table \ref{table:jobs_recommendation_ablation_models_}. 

